\documentclass{article}
\usepackage{microtype}
\usepackage{graphicx}
\usepackage{booktabs} 
\usepackage{amsmath}
\usepackage{mathtools}
\usepackage{bbm}

\usepackage{hyperref}
\usepackage{graphicx}

\usepackage[accepted]{icml2021} 

\usepackage{subfig}

\DeclareMathOperator*{\argmin}{{arg}\min}
\DeclareMathOperator*{\argmax}{{arg}\max}

\icmltitlerunning{Self-Supervised Neural Architecture Search for Imbalanced Datasets}

\begin{document}

\twocolumn[
\icmltitle{Self-Supervised Neural Architecture Search for \\ Imbalanced Datasets}

\begin{icmlauthorlist}
\icmlauthor{Aleksandr Timofeev}{lions}
\icmlauthor{Grigorios G. Chrysos}{lions}
\icmlauthor{Volkan Cevher}{lions}

\end{icmlauthorlist}

\icmlaffiliation{lions}{EPFL, Switzerland}

\icmlcorrespondingauthor{Aleksandr Timofeev}{aleksandr.timofeev@epfl.ch}

\icmlkeywords{Machine Learning, ICML}

\vskip 0.3in
]

\printAffiliationsAndNotice{} 
 
\begin{abstract}

Neural Architecture Search (NAS) provides state-of-the-art results when trained on well-curated datasets with annotated labels. However, annotating data or even having balanced number of samples can be a luxury for practitioners from different scientific fields, e.g., in the medical domain. To that end, we propose a NAS-based framework that bears the threefold contributions: (a) we focus on the self-supervised scenario, i.e., where no labels are required to determine the architecture, and (b) we assume the datasets are imbalanced, (c) we design each component to be able to run on a resource constrained setup, i.e., on a single GPU (e.g. Google Colab). Our components build on top of recent developments in  self-supervised learning~\citep{zbontar2021barlow}, self-supervised NAS~\citep{kaplan2020self} and extend them for the case of imbalanced datasets. We conduct experiments on an (artificially) imbalanced version of CIFAR-10 and we demonstrate our proposed method outperforms standard neural networks, while using $27\times$ less parameters. To validate our assumption on a naturally imbalanced dataset, we also conduct experiments on ChestMNIST and COVID-19 X-ray. The results demonstrate how the proposed method can be used in imbalanced datasets, while it can be fully run on a single GPU. Code is available \href{https://github.com/TimofeevAlex/ssnas_imbalanced}{here}.
\end{abstract}
\section{Introduction}
\label{sec:ssnas_imbalanced_intro}

Deep neural networks (DNNs) have demonstrated success in significant tasks, like image recognition~\citep{He_2016_CVPR} and text processing~\citep{devlin-etal-2019-bert}. Their stellar performance can be attributed to the following three pillars: a) well-curated datasets, b) tailored network architectures devised by experienced practitioners, c) specialized hardware, i.e. GPUs and TPUs. The adoption of DNNs by practitioners in different fields relies on a critical question: \\ \emph{Are those pillars still holding ground in real-world tasks?}

The first obstacle is that well-curated datasets (e.g., uniform number of samples over the classes) might be hard or even impossible to obtain in different fields, e.g., medical imaging. Similarly, when downloading images from the web, the amount of images of dogs/cats is much larger than the images of `Vaquita' fish. One mitigation of such imbalanced classes is the development of the Neural Architecture Search (NAS)~\citep{zoph2016neural}, which enabled researchers to build architectures that can then generalize to similar tasks. Obtaining the annotations required for NAS methods is both a laborious and costly process. To that end, self-supervised learning (SSL) has been proposed for extracting representations~\citep{bengio2013representation}. One major drawback of both NAS and SSL is that they require substantial computational resources, making their adoption harder. 

In this work, we propose a novel framework that combines NAS with self-supervised learning and handling imbalanced datasets. Our method relies on recent progress with self-supervised learning~\citep{zbontar2021barlow} and self-supervised NAS~\citep{kaplan2020self}. Specifically, the proposed method designs a network architecture using only unlabelled data, which are also (naturally) imbalanced, e.g., like data automatically downloaded from the web. We pay particular attention to the resources required, i.e., every component of the proposed framework is designed to run on a single GPU, e.g., on Google Colab. We evaluate our method using both a long-tailed distribution and naturally imbalanced datasets for medical imaging. In both settings, we notice that the proposed method results in accuracy that is similar to or better than well-established handcrafted DNNs with a fraction of the parameters of those networks, i.e., up to $27\times$ less parameters.

\section{Preliminaries}
\label{sec:background}
\subsection{Neural Architecture Search}
\label{sec:nas}

Neural Architecture Search (NAS) can be roughly separated into three components~\citep{elsken2019neural}: search space, search strategy, and performance estimation strategy. The first component defines the set of architectures to be explored; the second one determines how to explore the search space; the third one designs a way to estimate the performance in each step. The first approaches on NAS used evolutionary algorithms \citep{real2017large, real2019regularized} or reinforcement learning \citep{zoph2016neural} and outperform handcrafted neural architectures. One major drawback was the immense computational resources required for running NAS. The first papers that focus on the reduction of the computational cost construct a supernet that covers all possible operations and train exponentially many sub-networks simultaneously ~\citep{pham2018efficient, liu2018darts}. This approach indeed reduces the computational cost and is improved by the recent works. Specifically, \citep{cai2018proxylessnas} samples architecture paths during the search phase such that only one is trained each step. This allows training architecture of the same depth as the final model which eliminates the depth gap in performance. Similarly, \citep{Xu2020PC-DARTS} samples a small subset of channels and replace the rest with skip-connections. Both methods require less memory and reduce search time by using larger batches. However, it is shown that such approaches are unfair in the choice of operations which leads to deteriorating of sub-network performance \citep{chu2019fairnas, fairdarts} Another approach \citep{Liu_2020_CVPR} is based on growing and trimming candidate architectures which is combined with memory-efficient loss. 
In our work, we stick to the first approach. We avoid the depth gap using the same depth final networks and induce fair competition between operations leveraging \citep{fairdarts}.

The seminal work of \emph{DARTS}~\citep{liu2018darts} constructs a differentiable search space using a cell-based approach. The final architecture is constructed by stacking cells. Each cell is a directed acyclic graph (DAG) with $N$ nodes.  Each edge represents a candidate operation $o_{ij}$ with input $x_i$ and output $x_j$, where $x_j = \sum_{i < j} o_{ij}(x_i)$. This neural network is referred to a supernet or a parent network. Such a search space would be discrete, hence a softmax relaxation between the candidate operations $\mathcal{O} = \{o_{ij}^1, o_{ij}^2, ..., o_{ij}^M\}$ is used:
\begin{equation*}
    \overline{o}_{ij} = \sum_{o \in \mathcal{O}} \frac{\exp\left(\alpha_{o_{ij}}\right)}{\sum_{o^{'} \in \mathcal{O}} \exp\left(\alpha_{o_{ij}^{'}}\right)} o(x),
\label{eq:darts_softmax}
\end{equation*}
where $\alpha_{o_{ij}}$ is operation mixing weights. Then, NAS is reduced to learning these weights. The final discrete architecture is obtained by $o_{ij} = \argmax_{o \in \mathcal{O}} \alpha_{o_{ij}}$.
The following bi-level optimization problem describes the objective:
\begin{equation*}
    \min_{\alpha} \ell_{val}\left(\omega^{*}(\alpha), \alpha\right) 
    \mbox{ s.t. }\omega^{*}(\alpha) = \argmin_\omega \ell_{train}(\omega, \alpha).
\end{equation*}
where $\omega$ denotes normal neural network weights, $\mathcal{L}_{val}$ and $\mathcal{L}_{train}$ are loss functions computed based on batches from validation and train sets correspondingly.
It is hard to solve this task directly. Thus, we approximate $\omega^{*}(\alpha)$ using only a single training step
$
     \ell_{val}\left(\omega^{*}(\alpha), \alpha\right) \approx \ell_{val}\left(\omega - \xi \nabla_{\omega} \ell_{train}\left(\omega, \alpha\right), \alpha\right).
$

\paragraph{FairDARTS~\citep{fairdarts}} 

DARTS has the significant overhead of maintaining the supernet during training. To mitigate that, during the search phase a shallow network is assumed, which is then duplicated to obtain the full network for evaluation. However, the weights obtained by a shallow neural network are not appropriate for deep models \citep{9010638}. Specifically, skip connections are frequently selected as the operation $o_{ij}$. Additional drawback is lack of weights significantly outperforming others. The recent work of FairDARTS~\citep{fairdarts} mitigates those issues using the following two modifications: (a) it replaces the softmax operation with a sigmoid function to avoid the competition with skip connections as a candidate operation, (b) it encourages sparsity in the architecture weights by using the following zero-one loss: $\ell_{0-1} = - \frac{1}{N} \sum_{i=1}^N (\sigma(\alpha_i) - 0.5)^2$, where $\alpha_i,\quad 1 \leq i \leq N$, are architecture weights. The final loss for the architecture weights is $\ell_{total} = \ell_{val}(\omega^*(\alpha), \alpha) + \omega_{0-1} \ell_{0-1}$, where $\omega_{0-1}$ controls the strength of the zero-one loss.

\subsection{Barlow Twins}
\label{sec:barlow}
Supervised learning has demonstrated success in a number of domains, but requires a massive amount of annotations, while it ignores the enormous amount of unlabelled data that can provide complementary information. The effort to utilize unsupervised learning has been decades old process \citep{radford2015unsupervised, Doersch_2015_ICCV, bengio2013representation}, with the concept of self-supervised learning as a popular method of learning. The idea is to devise one task that the "target label" is known, and use losses developed for supervised learning. For instance, predicting the next word in a sentence enables utilizing the virtually unlimited text on the internet for unsupervised training; this is precisely the method used in the recent successful GPT \citep{radford2019language} and BERT models \citep{devlin-etal-2019-bert}. Similarly, in visual computing a host of tasks has been used for self-supervised learning~\cite{noroozi2016unsupervised, gidaris2018unsupervised, chen2020simple}.

By analogy to other successful self-supervised methods \citep{chen2020simple, chen2020exploring, caron2020unsupervised} Barlow Twins \citep{zbontar2021barlow} creates a pair of images for every original image. The pair is created by applying two randomly sampled transformations (e.g., random crop, horizontal flip, color distortion in pixels, etc). Similar pairs of images are created for every sample in the mini-batch. The model extracts the representations $z^A$ and $z^B$ of the two corresponding distorted versions of the original mini-batch. The idea is then to make the cross-correlation between $z^A$ and $z^B$ close to the identity. Specifically, the objective function is $\ell_{\mathcal{BT}} = \sum_{i} (1 - \mathcal{C}_{ii})^2 + \lambda  \sum_i \sum_{j \neq i} \mathcal{C}_{ij}^2,$ where $\lambda$ is a positive coefficient, $\mathcal{C}$ is a cross-correlation matrix of the outputs size computed between two outputs:
\begin{equation*}
    \mathcal{C}_{ij} = \frac{\sum_b z_{b,i}^A z_{b,j}^B}{\sqrt{\sum_b \left(z_{b,i}^A\right)^2}\sqrt{\sum_b \left(z_{b,j}^B\right)^2}},
\end{equation*}
where $b$ indexes batch samples and $i, j$ index the vector dimension of the outputs. In other words, the model is encouraged to differentiate the two distinct images in each pair. The advantages of Barlow Twins is that this decorrelation removes redundant information about samples in the output units. Unlike other recent self-supervised methods, Barlow Twins does not require large batches which is important when there are constrained resources (e.g., a single GPU).

\section{Methodology}
\label{sec:method}
We propose a new NAS-based approach for real-world datasets, which might not have available labels or they might be imbalanced. The approach consists of two steps: architecture search and subsequent fine-tuning.

Our method is build on top of FairDARTS. This allows eliminating the shortcomings of DARTS as mentioned in Section \ref{sec:nas}. We also replace the supervised loss with a self-supervised one. As \cite{NEURIPS2020_e025b627}  shows, it is also beneficial for learning imbalanced datasets. Namely, the recent Barlow Twins loss which does not require labels. Additionally, we use the supernet with only 3 cells for all steps. This is beneficial for three reasons. Firstly, the training process is efficient and affordable even for slow GPUs, while it produces small but powerful architectures. Secondly, the designed architecture is appropriate for the final model as its depth is unchanged. Thirdly, the learned weights in the first step are fully utilized in the second step (i.e., unlike other NAS methods that the weight values are typically discarded). To fine-tune the designed model, we add on the top another layer which projects the output matrix into the output classes and train it with the focal loss (see Appendix \ref{sec:handle_imbalance}) in a supervised manner. We do not freeze weights of the rest of the network. Furthermore, we apply the logit adjustment (see Appendix \ref{sec:handle_imbalance}) to improve learning of rare classes. The ablation study on imbalance handling techniques is in Appendix \ref{sec:appendix_d}.

Our work shares some similarities with the recent Self-Supervised Neural Architecture Search (SSNAS)~\citep{kaplan2020self}, which we describe next. SSNAS consists of three steps. In the first step, DARTS is used to determine a cell architecture by a shallow neural network. SSNAS assumes the unlabelled data and uses SimCLR \citep{chen2020simple} for determining the architecture. In the next step, SSNAS stacks the constructed cells from the previous step to obtain the architecture, which is sequentially trained with the same self-supervised loss. In the last step, the architecture is fine-tuned using an annotated dataset. Despite the similarities, our method differs from SSNAS in four critical ways: (a) Using DARTS has several drawbacks as aforementioned, while DARTS is not robust to initialization and it requires several runs, (b) SimCLR should be used with large batches, while Barlow Twins exhibits better performance and can be executed with a smaller batch size, (c) we use a smaller supernet which improves training time and produces more efficient architectures, (d) we skip the self-supervised pretraining step without occurring a loss in performance (this step required a $\approx 30\%$ overhead in the training time). Lastly, our method is specifically developed for imbalanced datasets by leveraging the logit adjustment and the focal loss.

\section{Experiments}
\label{sec:ssnas_imbalanced_experiments}

Below, we conduct an empirical validation of the proposed method in both a long-tailed distribution and a naturally imbalanced medical dataset. Furthermore, we validate whether the learned architecture can be used for transfer learning in the crucial domain of medical imaging, where we utilize a recent COVID-19 X-ray dataset. Our empirical evidence confirms that the proposed method can achieve similar performance with well-established networks using a fraction of their parameters. 

\paragraph{Training details.}  We apply a first-order version of FairDARTS to accelerate search. A supernet has 3 cells with 4 nodes each. The search space is the same as in \citep{liu2018darts, kaplan2020self}. 
We use SGD with learning rate $0.025$, momentum $0.9$, and weight decay $3 \times 10^{-4}$ with cosine annealing learning rate scheduler \citep{loshchilov2016sgdr}. A batch size of $32$ is used, while we train the architecture for $100$ epochs. The experiments are performed on NVIDIA Tesla K40c.
In fine-tuning step, the focal loss is applied with $\gamma=2$ and $\alpha_t = 1$. We use light data augmentation: random image cropping and horizontal flipping. The training is run for $600$ epochs or until convergence. The rest parameters are the same.

\subsection{Evaluation on a long-tailed distribution}
We evaluate the proposed method on the long-tailed version of CIFAR-10 dataset \citep{krizhevsky2009learning}. To this end, we reduce the number of samples for each class according to an exponential function $n^{lt}_{c} = \beta^{c + 1} n_{c}, \quad 0 \leq c < |C|,$ where $\beta \in (0, 1)$, $n_{c}$ and $n^{lt}_{c}$ are numbers of samples in each class before and after transformation correspondingly, $C$ is a set of classes. The test set stays unchanged. We define the imbalance factor $\rho$ of a dataset as the number of training samples in the largest class divided by the smallest one. The imbalanced dataset is used for both architecture search and subsequent fine-tuning. No label is used for the architecture search.

In Table \ref{tab:ssnas_experiments_cifar10}, we summarize the results of our experiments and compare them against the previous representative works on long-tailed distributions: ResNet-32 + Focal loss \citep{lin2017focal}, ResNet-32 + Sigmoid (SGM) and Balanced Sigmoid (BSGM) Cross Entropy losses \citep{cui2019class}, LDAM-DRW \citep{cao2019learning}, smDragon and VE2 + smDragon \citep{samuel2021generalized}, SSNAS \citep{kaplan2020self}. 
To provide a fair comparison to SSNAS method, we implement it in a common framework with common hyper-parameters. Notably, NAS-based methods require orders of magnitude less parameters. Though, SSNAS result in an increased error for the reduced parameters. 
Our method significantly improves the results of other methods achieving it with much fewer parameters.

\begin{table}
\caption{Image classification on CIFAR-10 LT ($\rho = 10$). The number of parameters is reported in millions ($\times 10^6$). }
\centering
    \vskip 0.1in
\scalebox{0.935}{
    \begin{tabular}{lccc}
    \toprule
    Method & \# Params & Error ($\downarrow$) \\
    \midrule
    ResNet-32 + Focal & 21.80 & 13.34 \\
    ResNet-32 + SGM   & 21.80 & 12.97\\
    ResNet-32 + BSGM   & 21.80 & 12.51\\
    LDAM-DRW & 21.80 & 11.84\\
    smDragon   & 21.80+ & 12.17\\
    VE2 + smDragon   & 21.80+ & 11.84\\
    \midrule
    SSNAS~\citep{kaplan2020self}      & 0.83 & 18.84 \\
    Our method    & \textbf{0.81} &  \textbf{10.91} \\
    \bottomrule
    \end{tabular}
    }
    \vskip -0.1in
\label{tab:ssnas_experiments_cifar10}
\end{table}

\subsection{Evaluation on naturally imbalanced datasets}
We assess the performance of the method on \emph{ChestMNIST}  \citep{chest} which is a naturally imbalanced dataset. 
The dataset contains 78,468 images of chest X-ray scans. There are 14 non-exclusive pathologies. The results are presented in Table \ref{tab:ssnas_experiments_chest}. The accuracy of models for comparison (ResNet \citep{He_2016_CVPR}, auto-sklearn \citep{feurer2019auto}, AutoKeras \citep{jin2019auto}, and Google Auto ML) are reported from \citep{medmnist}. The proposed method is able to achieve the best seen before result wherein it keeps a tiny number of parameters. The smaller number of parameters for SSNAS is caused by a lot of skip-connections. 

\begin{table}
\caption{Image classification on ChestMNIST. The number of parameters is reported in millions ($\times 10^6$). The resolution of input images are indicated in the parenthesis.}
\vskip 0.1in
\begin{center}
\begin{tabular}{lccc}
\toprule
Method & \# Params & Accuracy ($\uparrow$)\\
\midrule
ResNet-18 (28)  & 11.68 & 94.7 \\
ResNet-18 (224) & 11.68 & \textbf{94.8} \\
ResNet-50 (28)  & 25.56 & 94.7 \\
ResNet-50 (224) & 25.56 & 94.7 \\
auto-sklearn (28)    &   -   & 64.7 \\
AutoKeras (28)       &   -   & 93.9\\
Google Auto ML (28)  &   -   & 94.7 \\
\midrule
SSNAS  (28)    & \textbf{0.57} &  94.7 \\
Our method (28)    & 0.82 & \textbf{94.8} \\
\bottomrule
\end{tabular}
\end{center}
\vskip -0.1in
\label{tab:ssnas_experiments_chest}
\end{table}

As typically done in NAS, we evaluate the optimized architecture on transfer learning using \emph{COVID-19 X-ray} dataset \citep{OZTURK2020103792}. This dataset consists of chest images and naturally imbalanced with $\rho=4$. This dataset represents several difficulties that arise in real-world settings: (a) there is an imbalance factor $\rho=4$, (b) the images are slightly different since they are collected from different sources, (c) there is noise in some images since there are some overlaid labels which are not related to the task in hand (see Appendix \ref{sec:appendix_a}).  
Unfortunately, because such datasets are recent, we have the only model to compare which is DarkCovidNet \citep{OZTURK2020103792}. Table \ref{tab:ssnas_experiments_covid} shows that our architecture is successfully transferred to another task achieving slightly better results DarkCovidNet but with smaller resolution and number of parameters.

\begin{table}
\caption{Image classification on COVID-19 X-ray. The number of parameters is reported in millions ($\times 10^6$). The resolution of input images are indicated in the parenthesis.} 
\vskip 0.1in
\begin{center}
\begin{tabular}{lccc}
\toprule
Method & \# Params & Accuracy ($\uparrow$) \\
\midrule
DarkCovidNet (224)  & 1.16 & 98.08 \\
\midrule
SSNAS (224)     & \textbf{0.57} &  \textbf{98.40}   \\
SSNAS  (28)    & \textbf{0.57} &  98.08   \\
Our method (224)   & 0.82 & \textbf{98.40} \\
Our method (28)  & 0.82 & \textbf{98.40} \\
\bottomrule
\end{tabular}
\end{center}
\vskip -0.1in
\label{tab:ssnas_experiments_covid}
\end{table}

\section{Conclusion}
In this paper, we propose a NAS framework which is well-suited for scenarios with real-world tasks, where the data are naturally imbalanced and do not have label annotations. Our framework designs an architecture based on the provided unlabelled data using self-supervised learning. To evaluate our method, we conduct experiments on a long-tailed version of CIFAR-10 as well as ChestMNIST and COVID-19 X-ray which are medical datasets that are naturally imbalanced. For all the experiments, we show that the proposed approach provides more compact architecture while maintaining an accuracy on par with strong performing baselines. We expect our method to provide a reasonable framework for practitioners from different fields that want to capitalize on the success of deep neural networks but do not necessarily have well-curated datasets. In addition, our method is suitable for researchers on a constrained budget (e.g., using only the publicly-available Google Colab). 
\section{Acknowledgements}
\label{sec:ssnas_acks}

This project was sponsored by the Department of the Navy, Office of Naval Research(ONR) under a grant number N62909-17-1-2111, by Hasler Foundation Program: Cyber Human Systems (project number 16066). This project has received funding from the European Research Council (ERC) under the European Union's Horizon 2020 research and innovation programme (grant agreement number 725594). The project was also supported by 2019 Google Faculty Research Award.

\bibliography{main.bib}
\bibliographystyle{icml2021}
\appendix
\section{Handling imbalanced datasets}
\label{sec:handle_imbalance}
\setcounter{figure}{0}  
\setcounter{table}{0} 
\renewcommand\thefigure{A\arabic{figure}}  
\renewcommand\thetable{A\arabic{table}}
\paragraph{Logit adjustment.} In many real-world applications, gathering balanced datasets is difficult or even impossible. For instance, in medical analysis, a small group of the patients has a specific pathology that the majority of the population does not have. In such settings, doing predictions on imbalanced datasets is crucial. 

In \citep{menon2020long}, the authors propose the logit adjustment for the loss function to handle imbalance which corrects the output of the model before softmax operation. Specifically, they introduce the logit adjusted softmax cross-entropy loss:
\begin{equation*}
    \ell\left(y, f(x)\right) = - \log \frac{e^{f_y(x) + \tau\log\pi_y}}{\sum_{y^{'}\in [L]} e^{f_{y^{'}}(x) + \tau\log\pi_{y^{'}}}},
\end{equation*}
where $L$ is a number of classes, $f_y(x)$ is a logit of the given class, $\pi_y$ is empirical frequencies of classes. Therefore, we induce the label-dependent prior offset which requires a larger margin for rare classes. 

\paragraph{Focal loss.} The focal loss~\citep{lin2017focal} is frequently used in imbalanced datasets \citep{sambasivam2021predictive, dong2021recognition}. The idea behind the focal loss is to give a lower weight to easily classified samples.In a binary case, we introduce $p_t = \mathbbm{1}[y = 1]p + \mathbbm{1}[y = -1](1-p)$,where  $y \in \{-1, 1\}$ are labels, $p$ is model's estimated probability, and $\mathbbm{1}[\cdot]$ is an indicator function. The cross-entropy loss is then $\mbox{CE}(p_t) = - \log p_t$. To tackle the imbalance problem, the focal loss adds a modulating factor to the weighted cross-entropy $\mbox{FL}(p_t) = -\alpha_t (1-p_t)^\gamma \log p_t,$ where $\alpha_t$ are loss weights which can be inverse class frequencies,  $\gamma$ is a tunable parameter. If a sample is misclassified and $p_t$ is small, the loss is unaffected. However, when $p_t \to 1$ then  $(1-p_t)^\gamma \to 0$ which gives less weight to this sample.

\section{Ablation study}
\label{sec:appendix_d}
\setcounter{figure}{0}  
\setcounter{table}{0} 
\renewcommand\thefigure{B\arabic{figure}}  
\renewcommand\thetable{B\arabic{table}}
To show effectiveness of components responsible for handling imbalance, we analyse the performance of all combinations of the Focal loss and the logit adjustment as well as their absence. In the latter case, we use simply the cross-entropy loss. The results are summarized in Table \ref{tab:ablation}. The best result is achieved by combination of the focal loss and logit adjustment. Removing the latter slightly deteriorates the performance while abscence of the focal loss is significant. 

\begin{table}
\caption{Ablation study on imbalance handling techniques for image classification on CIFAR-10 LT ($\rho = 10$). CE = Cross-Entropy loss, FL = Focal loss.} 
\vskip 0.1in
\begin{center}
\begin{tabular}{lcc}
\toprule
Method & Error ($\downarrow$) \\
\midrule

CE  &              13.21 \\
CE + Logit adj. &  13.05 \\
FL &               11.78 \\
FL + Logit adj. &  10.91 \\
\bottomrule
\end{tabular}
\end{center}
\vskip -0.1in
\label{tab:ablation}
\end{table}

\section{COVID-19 X-ray dataset}
\label{sec:appendix_a}
\setcounter{figure}{0}  
\setcounter{table}{0} 
\renewcommand\thefigure{C\arabic{figure}}  
\renewcommand\thetable{C\arabic{table}}

\begin{figure}[ht]
\vskip 0.2in
\begin{center}
    \subfloat{\includegraphics[width=0.24\textwidth]{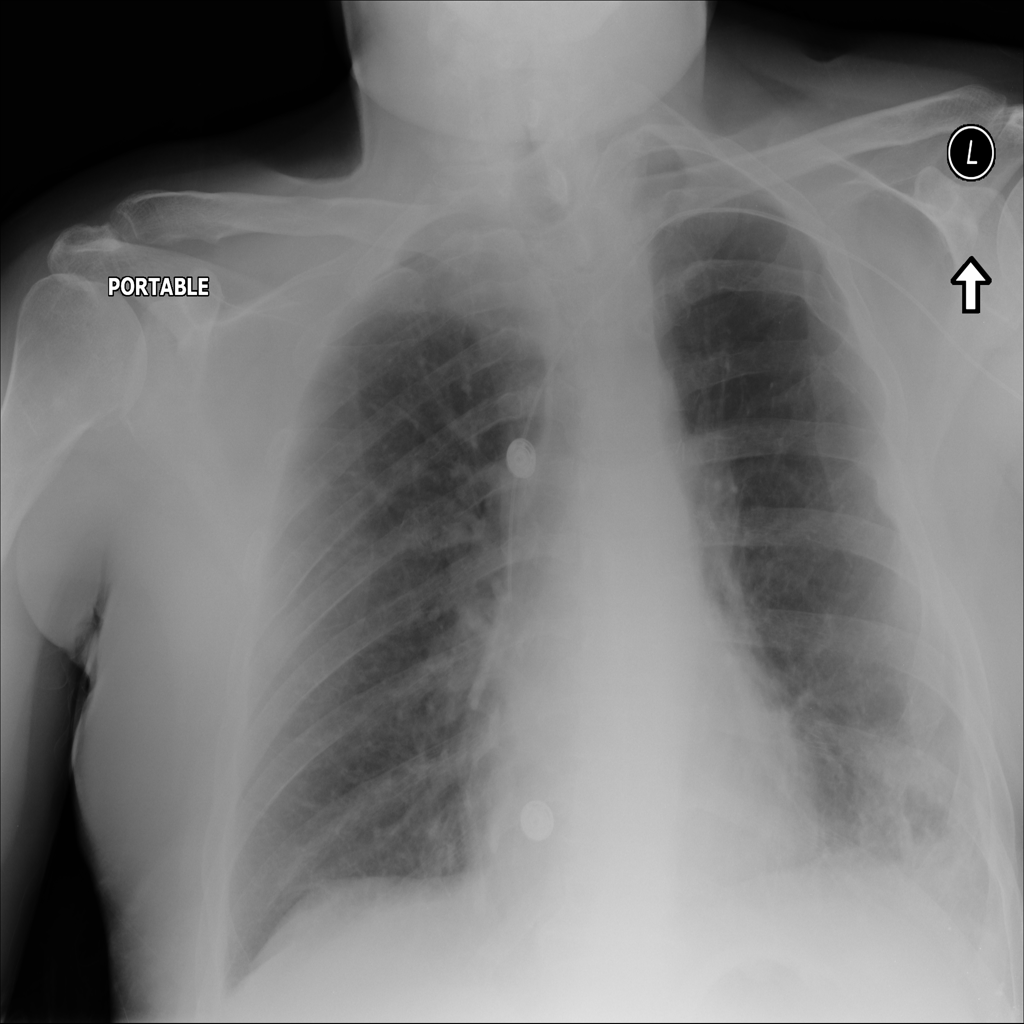}\hspace{2pt}}
    \subfloat{\includegraphics[width=0.24\textwidth]{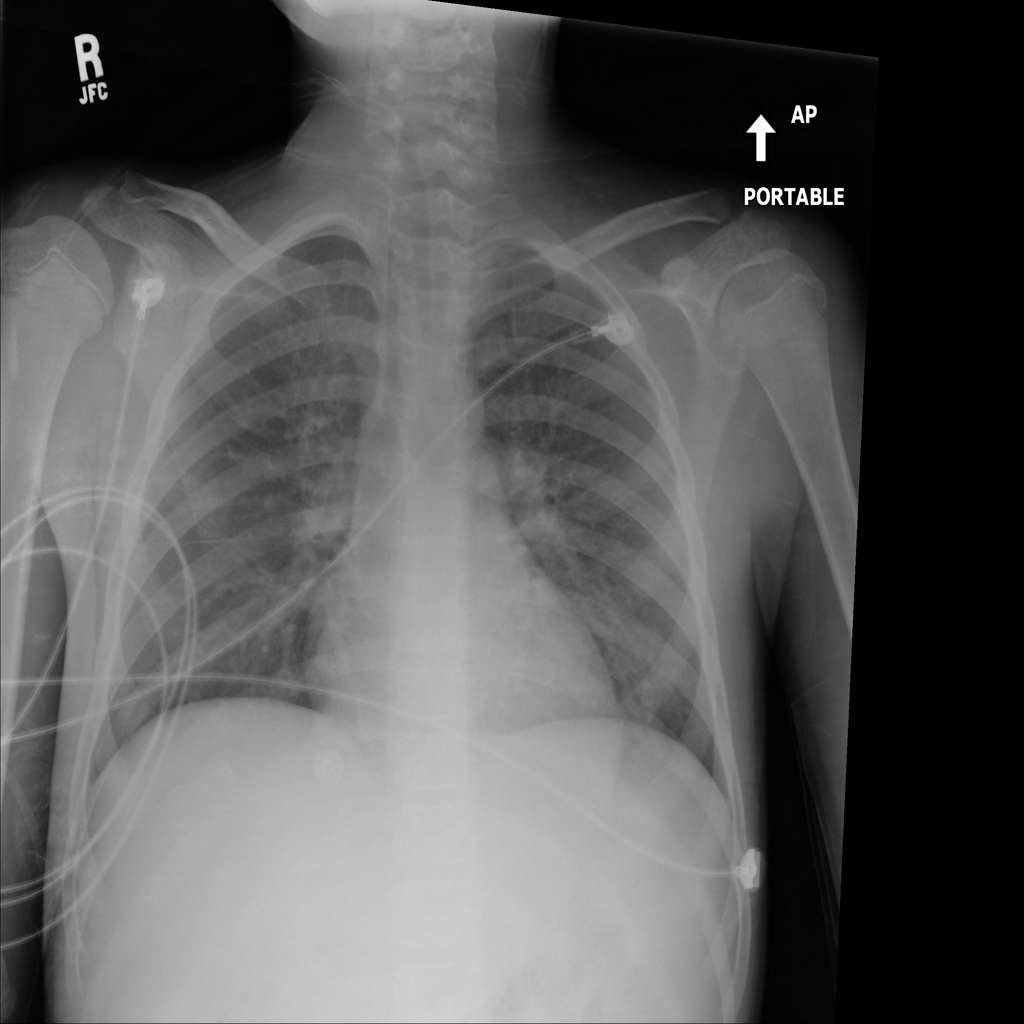}} \\
    \subfloat{\includegraphics[width=0.24\textwidth]{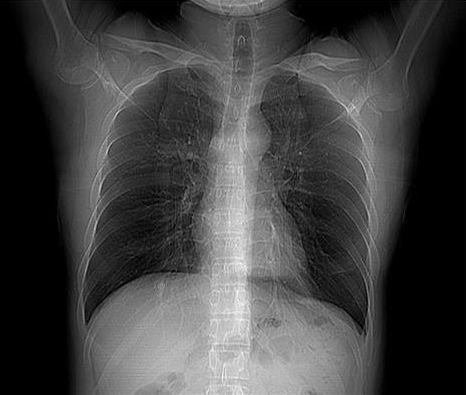}\hspace{2pt}}
    \subfloat{\includegraphics[width=0.205\textwidth]{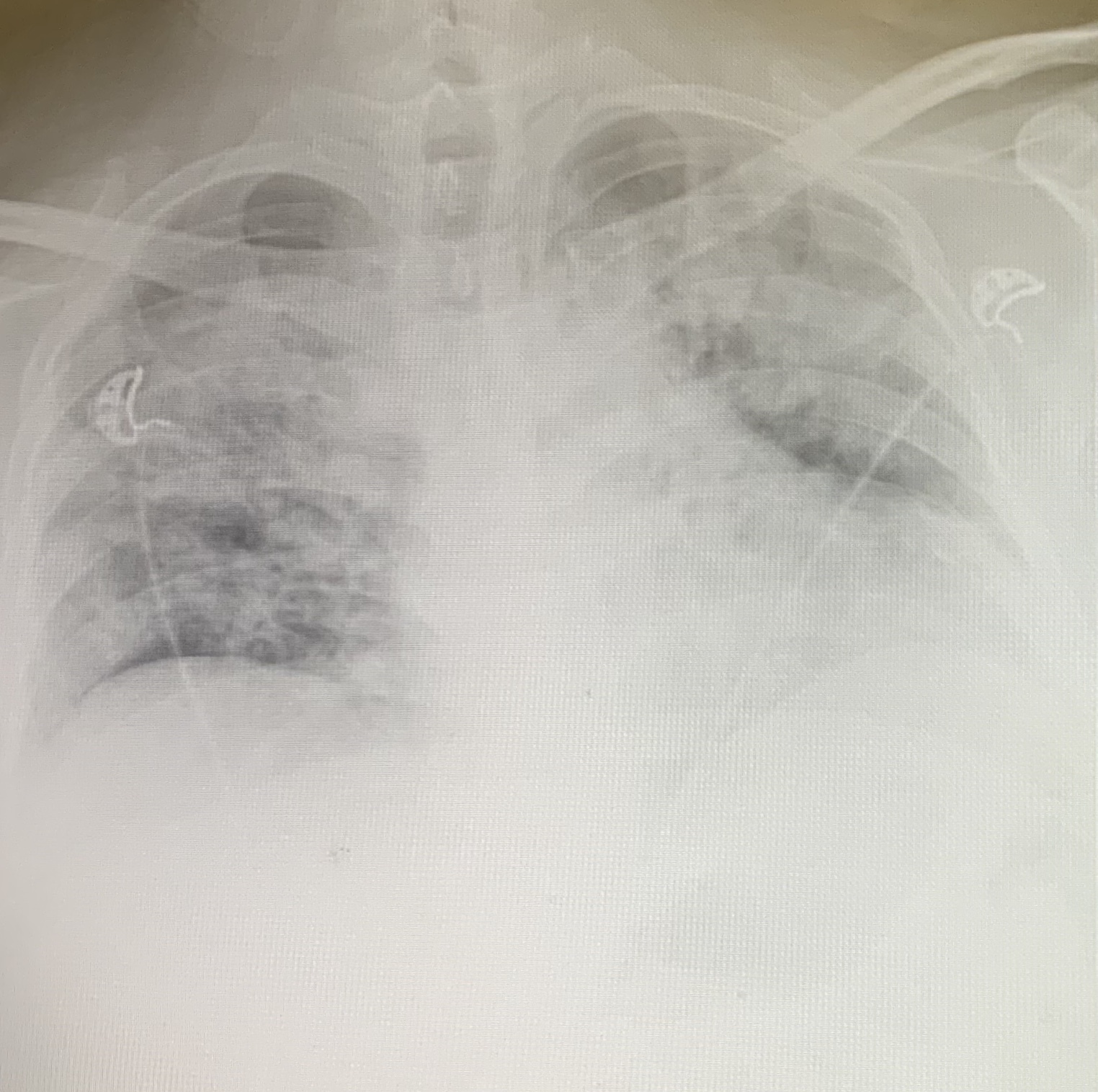}}
\caption{The samples of COVID-19 X-ray dataset. \textit{Top row:} no findings. \textit{Bottom row:} COVID-19.}
\label{fig:covid19}
\end{center}
\vskip -0.2in
\end{figure}

In Figure \ref{fig:covid19}, we show four representative samples of the COVID-19 X-ray dataset. All images are collected from different sources, while some images contain unrelated content overlaid on image. Likewise, the light intensity, the resolutions, and the image formats might differ from image to image which makes learning harder.

\end{document}